\pdfoutput=1
\documentclass[11pt,a4paper]{article}
\usepackage[hyperref]{eacl2021}
\usepackage{times}
\usepackage{latexsym}

\usepackage{comment}

\usepackage{microtype}

\aclfinalcopy % Uncomment this line for the final submission
\title{How Vulnerable Are Automatic Fake News Detection Methods to Adversarial Attacks?}

\author{Camille Koenders \\
  Technische Universität Berlin \\
  \texttt{camille.koenders@gmail.com} \\\And
  Johannes Filla \\
  Technische Universität Berlin \\
  \texttt{johannesfilla@gmx.de} \\\AND
  Nicolai Schneider \\
  Technische Universität Berlin \\
  \texttt{nicolai-schneider@t-online.de} \\\And
  Vinicius Woloszyn \\
  Technische Universität Berlin \\
  \texttt{woloszyn@tu-berlin.de}\\
  }

\date{}

\begin{document}
\maketitle
\begin{abstract}
As the spread of false information on the internet has increased dramatically in recent years, more and more attention is being paid to automated fake news detection. Some fake news detection methods are already quite successful. Nevertheless, there are still many vulnerabilities in the detection algorithms. The reason for this is that fake news publishers can structure and formulate their texts in such a way that a detection algorithm does not expose this text as fake news. This paper shows that it is possible to automatically attack state-of-the-art models that have been trained to detect Fake News, making these vulnerable. For this purpose, corresponding models were first trained based on a dataset. Then, using Text-Attack, an attempt was made to manipulate the trained models in such a way that previously correctly identified fake news was classified as true news. The results show that it is possible to automatically bypass Fake News detection mechanisms, leading to implications concerning existing policy initiatives. 

% As the spread of false information on the internet has increased dramatically in recent years, more and more attention is being paid to automated fake news detection. Some fake news detection methods are already quite successful. Nevertheless, there are still many vulnerabilities in the detection algorithms. The reason for this is that fake news publishers can structure and formulate their texts in such a way that a detection algorithm does not expose this text as fake news. This paper aims to show the vulnerability of state-of-the-art models that have been trained to detect Fake News. For this purpose, corresponding models are first trained on the basis of a dataset. Then, using Text-Attack, an attempt is made to manipulate the trained models in such a way that previously correctly identified fake news is classified as true news. Additionally, this paper addresses the consequences formulated from a political and technological perspective to reduce fake news on the Internet and discusses the impact of AI-disinformation initiatives on freedom of expression, media pluralism and democracy.

\end{abstract}

%% CHAPTERS

% INTRODUCTION
\section{Introduction}

The spreading of disinformation throughout the web has become a serious problem for a democratic society. The dissemination of fake news has become a profitable business and a common practice among politicians and content producers. A recent study entitled ``Regulating disinformation with artificial intelligence'' \cite{euregulation}, examines the trade-offs involved in using automated technology to limit the spread of disinformation online. Based on this study, this paper discusses the social and technical problems of Automatic Content Moderation (ACM) poses to freedom of expression.
\\

Although AI and Natural Language Generation have evolved tremendously in the last decade, there are still concerns regarding the potential implications of automatically using AI to moderate content. One problem is that automatic moderation of content on social networks will accelerate a race in which AI will be created to counter-attack AI. 
\\
Adversarial machine learning is an technique that attempts to fool models by exploiting vulnerabilities and compromising the results. For example, by changing particular words - e.g., from ``Barack" (Obama) to ``b4r4ck" - it is possible to mislead classifiers and overpass automatic detection filters. Recent works \cite{zhou2019fake} show the state-of-the-art machine learning models are vulnerable to these attacks. 

This study relies on state-of-the-art techniques to attack and dive deep into the fake news detection vulnerabilities. The goal is to experiment with adversarial attacks to discover and compute the vulnerabilities of fake news classifiers. Therefore, this work aims to answer the following research question: \emph{How vulnerable is fake news detection to adversarial attacks?}

The remainder of this paper is organized as follows. Section \ref{sec:relatedwork} discusses the background and previous related works. Section \ref{sec:experimentdesign} describes the design of our experiments, and Section \ref{sec:results} presents the results and discussion. Section \ref{sec:conclusion} summarizes our conclusions and presents future research directions.

% RELATED WORK
\section{Related Work}
\label{sec:relatedwork}

\subsection{Fake News Detection}
Fake news are used to manipulate general opinions of readers about a certain topic \cite{surveyzhouzafarani}. Unlike typical ``clickbait" articles, which use misleading and eye-catching headlines, fake news are usually quite long and wordy, consisting of inaccurate or invented plots \cite{chakraborty2016stop}. This gives rise to the assumption of a well-researched and factually correct article. The reader, thus, does not notice how their personal opinion about a certain topic is deliberately manipulated.

Fake news detection refers to any kind of identification of such fake news. Due to the speed at which digital news is produced today, effective, automated fake news detection requires the use of machine learning tools. Previous research has mainly focused on fake news in social media and fake news in online news articles \cite{q1_fakeflow}. There are various models of the machine detection of fake news, which are based on different heuristics.

For example, Ghanem et al. \cite{q1_fakeflow} discuss the effectiveness of the ``FakeFlow" model, which incorporates both word embedding and affective information such as emotions, moods or hyperbolic words, based on four different datasets. The model receives several small text segments as input instead of an entire article. The result of the study was that this model is more effective than most state-of-the-art models. It generated similar results with less resources.

Another study \cite{q2_claimreview} investigates the possibility of using artificial intelligence for the automatic generation of Claim-Review. Claim-Review is the web markup introduced in 2015 that allows search engines to access fact-checked articles. The basic idea of the so-called ``fact check" is that journalists and fact checkers identify misinformation and prevent it from spreading. Accordingly, it is important that fact-checked articles are highlighted and shared by users.

Furthermore, research is currently looking at when and why a news article is identified as fake news and when it is not \cite{q3_defend}. This research on ``explainable fake news detection" aims to improve the detection performance of the algorithms. For this purpose, both news content and user comments are used as data input.

\subsection{Adversarial Attack}
Adversarial attacks are part of adversarial machine learning, which has become increasingly important in the field of applied artificial intelligence in recent years. In an adversarial attack, the input data of a neural network is intentionally manipulated to test how resistant it is to deliver the same outputs. These manipulated input data is called ``adversarial examples". \cite{via_NLP} Such a neural network is described as a ``fake news detector" in the context of this paper. The reason for research in the field of adversarial attacks is that more and more attempts are being made to outsmart fake news detectors on the internet. This can be done, for example, by changing the spelling of a word so that the word remains easily interpretable for humans but not for an algorithm (from ``Barack" to ``B4r4ck"). For this reason, the input data is always manipulated in such a way that a human hardly notices any differences from the non-manipulated input data. In the area of object recognition, for example, only individual pixels are slightly changed, which the human eye can hardly perceive. 

Examples of adversarial examples in the field of fake news detection are:
\begin{itemize}
 \item \textbf{Fact distortion:} Here, some words are changed or exaggerated. It can be about people, time or places.
 \item \textbf{Subject-object-exchange:} By exchanging subject and object in a sentence, the reader is confused as to who is the executor and who is the recipient.
 \item \textbf{Cause confounding:} Here, either false causal connections are made between events or certain passages of a text are omitted.
\end{itemize}

Up to now, vulnerabilities to adversarial attacks have been identified in all application areas of neural networks. Especially in the context of increasingly safety-critical tasks for neural networks (e.g., autonomous driving), methods for detecting false or distorted input data are becoming more and more relevant. \cite{textattack}.

%\subsection{Legal framework on AI}

\subsection{Regulating disinformation with artificial intelligence}
According to the study ``Regulating disinformation with artificial intelligence" \cite{euregulation}, disinformation is defined as 'false, inaccurate, or misleading information designed, presented and promoted to intentionally cause public harm or for profit. This definition is based on the definition of ``Final report of the High Level Expert Group on Fake News and Online Disinformation" \cite{expert_group}, which additionally specifies that the term ``disinformation" does not include fundamentally illegal content such as hate speech or incitement to violence. Nor does it include misinformation that is clearly not misleading, such as satire or parody. As a further delimitation, the paper defines the term ``misinformation", which refers to any misinformation that is unintentionally or accidentally false or inaccurate. 

Marsden and Meyer explain the causes of disinformation in the online context and the responses that have been formulated from a technological perspective. Furthermore they analyze the impact of AI-disinformation initiatives on freedom of expression, media pluralism and democracy. 

The issue of disinformation is a very long-term historical problem with human society. It has just got a global effect through automation, through the internet and new technologies. Even in the past, people faced the challenges of filtering out false information or illegal content from newly emerging media such as newspapers, radio or television. So, on the one hand there is currently a desire within the European Commission to take action against illegal and unwanted content through online intermediaries. On the other hand, the intervention and regulation of content on the internet is also seen critically, because this is against the basic concept of the internet with its freedom of expression. However, to stop the global spread of disinformation, a restriction of freedom of expression is necessary. Accordingly, for measures to filter fake news, there are three principles that must exist when restricting freedom of expression \cite{euregulation}:  
\begin{itemize}
 \item Measures must be established by law
 \item Measures must be legitimate and shown to be necessary
 \item Measures must be the least restrictive method of pursuing the goal
\end{itemize}

There are two different ways to detect and remove disinformation: The human and the technical moderation by AI. Over time, AI solutions have become increasingly effective in detecting and removing illegal or unwanted content, but they also raise the question of who is the judge in deciding what is legal or illegal and what is wanted or unwanted in society. The problem here is that neither law nor technology can be truly neutral. Both reflect the values and priorities of those who designed them. This principal is called “garbage in – garbage out”. According to expert opinions, AI can help to make an initial filtering, especially for texts and articles, which is then checked by humans. If AI is wrong, there is always the possibility to reverse the decision. To do this, companies also hire cheap subcontractors in remote countries to remove content.

Policy initiatives in the past have focused on making internet intermediaries more responsible for reducing disinformation on their platforms. While actual content creators were responsible for their content in the past, now the platforms have to take more and more responsibility for the content of the individual actors. The reactions of intermediary sites such as Facebook and YouTube are technological initiatives that identify certain content and remove it in different ways or do not publish it at all. The three most popular initiatives are filtering, blocking, deprioritisation:

\textbf{Filtering} is the most effective method. There are 2 different types of filtering: ex-ante and ex-post. This means filtering before a content goes online and filtering after it has already been published. With the exception of obviously illegal content (e.g. child abuse images or terrorist content) or disruptive content (e.g. spam, viruses), the ex post-removal is preferable. The reason for this is that there is better legal justification for this. One example of filtering is YouTube Content ID. With YouTube Content ID, uploaded files are matched against databases of works provided by copyright holders. If a match is found, copyright owners can decide whether to block, monetise or track a video containing their work.

\textbf{Blocking} is probably the most widespread method. It is used by users, email providers, search engines, social media platforms and network and internet access providers. Similar to filtering, blocking can take place ex-ante or ex-post, i.e. after knowledge, request or order. Blocking means that the content is not completely removed but blocks certain users from accessing the content. For this reason, it provides one significant advantage: Content can be blocked depending on the provider's terms of use or the laws of a particular region. For example, content may be blocked in certain places but may be available in others, e.g. if some countries allow certain content by law while others do not.

The last option is \textbf{deprioritisation}. In the context of disinformation, deprioritisation means that content is de-emphasized in users' feeds. This takes place when correct content from certain providers is displayed side by side with incorrect content. Then the wrong information is identified and deprioritised so that it is displayed further down and not so prominently anymore.

% DATASET
\section{Data Set}
\label{sec:dataset}
The data used in this paper was collected by a project called Untrue.News: A search engine designed to find fake stories on the internet \cite{woloszyn2020untrue}. It uses an open-source web crawler for searching fake news. This web crawler is connected to a natural language processing pipeline that uses automatic and semi-automatic strategies for data enrichment. In this way approximately 30.000 documents have been retrieved, dating back to the year 1995 (English documents). For our research we used 25.886 English sentences of these documents.  More languages, that were not used for our project can be found at untrue.news. \citet{woloszyn2020untrue} use four categories to classify their documents:

\begin{itemize}
    \item \textbf{TRUE}: completely accurate statements
    \item \textbf{FALSE}: completely false statement
    \item \textbf{MIXED}: partially accurate statements with some elements of falsity 
    \item \textbf{OTHER}: special articles that do not provide a clear verdict or do not match any other categories
\end{itemize}

These can be found in \cite{tchechmedjiev2019claimskg} and are less than those found in the schema.org/ClaimReview markup. For performing the text attacks, we only used the TRUE and FALSE statements. In Table \ref{table:multiclass} classification results can be found for the models trained with all four categories.

% EXPERIMENT DESIGN
\section{Experiment Design}
\label{sec:experimentdesign}

To understand the vulnerability of models trained to detect fake news we split our experiment into two steps. First training state-of-the-art machine learning models using the dataset described in Section \ref{sec:dataset} and second applying adverserial attacks on the dataset using TextAttack to manipulate the trained models to classify fake news as TRUE news.

\subsection{Fake News Detection as a Classification Problem}
Two types of classifiers were used. \textbf{Bin} a binary classifier, in which a positive class represents true news, and the negative not being true news and \textbf{Mult} a multiclass classifier, where each document is classified as either being true news, fake news, untrue news or other.

\subsection{Pre-trained Models}
In this subsection, we present the applied pre-trained models and BERT \cite{bert} which two of the models are based on. In this context a token describes a single word of a given text and a segments describes a sequence of tokens. The trained models will later be attacked by the Text Attack recipes described in Section \ref{sec:textattack}.

\subsubsection{BERT}
The BERT (Bidirectional Encoder Representations from Transformers) model was originally trained on two datasets: \textit{BookCoorpus} \cite{bookcorpus} and \textit{English Wikipedia}.
\\
The way BERT analyzes the text is by taking concatenations of two segments. The total length of the concatenation is bound by a parameter $T$ and is computed as one input with particular tokens in between describing the \textit{beginning} and \textit{end} of the sentence, as well as the separation point of the two segments. 
\\
During pretraining, BERT applies a \textit{Masked Language Model} (MLM) and a \textit{Next Sentence Prediction} (NSP). For MLM BERT selects 12\% of input tokens and replaces them with a \textit{[MASK]} token and another 1.5\% with a random vocabulary token. It then proceeds with predicting the selected \textit{[MASK]} tokens. With NSP BERT makes a binary prediction on whether two segments in a text are adjacent.

\subsubsection{RoBERTa}
The pre-trained RoBERTa (Robustly optimized BERT approach) \cite{roberta} is an optimized version of the original BERT model. The authors use a total length of concatenation $T=512$ (-longer than previously used) and expand the used datasets to: \textit{BookCoorpus}, \textit{CC-News} \cite{ccnews}, \textit{OpenWebText} \cite{openwebtext} and \textit{Stories} \cite{storiesdataset}. Furthermore, RoBERTa trains using a bigger batch size, removes NSP and changes the the masking for MLM each time a new input is given to the model therefore avoiding constant masking during different training epochs.

\subsubsection{BERTweet}
BERTweet \cite{bertweet} is a pre-trained model that uses elements from both BERT and RoBERTa. For instance, it's architecture is taken from BERT, while the pre-training procedure is "copied" from RoBERTa. The main difference is the used dataset: the only used dataset consists of 850 English tweets together accumulating 80GB of memory.

\subsubsection{Flair Embeddings}
The main difference of Flair Embeddings \cite{flairembeddings} to other models is their capturing of words which considers token as a sequence of characters. Furthermore, Flair Embeddings, also called contextual string embeddings, take contextual words into consideration, i.e. words that would appear consequently or previously. A word would therefore be embedded depending on the sequence of words. For our training we use the embeddings \textit{news-forward} and \textit{news-backward}, which were both trained on a 1 billion word corpus.

\subsection{Parameterization}
The training of each model runs for $E=15$ Epochs each, since at the small amount of training samples all models start converging after approximately 7-8 epochs. The mini batch size is set to $b=32$. For BERTweet and RoBERTa we specify the learning rate to $lr=3-e5$, while for Flair Embeddings to $lr=0.1$. Finally, for Flair Embeddings we also set the anneal factor (describing the factor by which the learning rate is annealed) to $a_f=0.5$ and the patience (the number of epochs without improvement until the learning rate would be annealed) to $p=5$. \noindent $tp$ is the number of positive instances correctly classified as positive, $tn$ number of negative instances correctly classified as negative, $fp$ negative instances wrongly classified as positive, and $fn$ is the number of positive instances wrongly classified as negative. We defined positive instances as fake news websites and negative instances as reliable news websites.

\subsection{Text Attack Recipes}
\label{sec:textattack}
The following attack recipes from TextAttack \cite{textattack} were applied on the dataset (described in Section \ref{sec:dataset}) to manipulate the three introduced models to classify FALSE statements as TRUE, therefore misinterpreting fake news as true news.

\begin{itemize}
    \item \textbf{DeepWordBug}: Generates small text perturbations in a black-box setting. It uses different types of character swaps (swapping, substituting, deleting and inserting) with greedy replace-1 scoring \cite{gao2018blackbox}.
    \item \textbf{Pruthi}: Simulates common typos with a concentration on QWERTY keyboard. Uses character insertion, deletion and swapping \cite{pruthi2019combating}.
    \item \textbf{TextBugger}: These attacks were optimized to perform with real world applications. They use space insertions, character deletion and swapping. Additionally they substitute characters with similar looking letters (ex. o with 0) and replace words with top nearest neighbor in context-aware word vector space \cite{Li_2019}.
    \item \textbf{PSOZang}: Word level attack using a sememe-based word substitution strategy as well as particle swarm optimization \cite{Zang_2020}.
    \item \textbf{PWWSRen2019}: These attacks focus on maintaining lexical correctness, grammatical correctness as well as semantic similarity by using synonym swap. Words for swap are prioritized on a combination of there saliency score and maximum word-swap effectiveness \cite{ren-etal-2019-generating}.
    \item \textbf{TextFoolerJin2019}: Word swap with their 50 closest embedding nearest neighbors. Optimized on BERT \cite{jin2020bert}.
    \item \textbf{IGAWang}: Implemented as an adversarial attack defense method. Uses counter-fitted word embedding swap \cite{wang2021natural}.
    \item \textbf{BAEGarg2019}: Uses a BERT masked language model transformation. It uses the language model for token replacement to best fit overall context \cite{garg2020bae}.
    \item \textbf{CheckList2020}: Inspired by the principles of behavioral testing. Uses changes in names, numbers and locations as well as contraction and extension \cite{ribeiro2020accuracy}.
    \item \textbf{InputReductionFeng}: This attack concentrates on the least important words in a sentence. It iteratively removes the world with the lowest importance value until the model changes its prediction. The importance is measured by looking at the change in confidence of the original prediction when removing the word from the original sentence \cite{feng2018pathologies}
\end{itemize}

With exception of \textit{InputReductionFeng} each recipe has three possible results for the attack on each sentence. \textit{Success} means the text attack resulted in a wrong classification. \textit{Skipped} mean that the model classified the sentence wrongly to begin with, therefore the sentence doesn`t need to be manipulated. \textit{Fail} means the model still classified the sentence correctly. \textit{InputReductionFeng} uses \textit{Maximized} to indicate that the model uncertainty was maximized. A rubbish example is classified as correct with a higher accuracy than the original valid input. \textit{Skipped} is used when the model classified the sentence wrongly to begin with, therefore the sentence doesn`t need to be manipulated.

% RESULTS AND DISCUSSION
\section{Results and Discussion}
\label{sec:results}
The results discussed in this section can also be found in our Github repository \cite{githubcode} including the respective implementations.

\subsection{Training the Models}
The first step of our experiment is to train state-of-the-art machine learning models to detect fake news. Since our dataset includes classes beyond TRUE and FALSE, we trained RoBERTa, BERTweet and Flair Embeddings both as binary and multi class classifier.

\subsubsection{Binary Classification}
\label{sec:binclass}
For the binary classification RoBERTa, BERTweet and FlairEmbeddings were only trained on TRUE and FALSE statements. Their precision, recall and F1-scores, depicted in Table \ref{table:binclass}, demonstrate that the BERT based models performed best with scores of $>80\%$. Overall, BERTweet had the best results, however, the difference to RoBERTa is minimal even though the models were pre-trained on entirely different datasets. With an F1-score of ~70\% FlairEmbeddings received the worst score which could be traced back to the small size of the dataset.
\begin{table}[ht!]
\centering
\begin{tabular}{llll}
\hline \textbf{Model} & \textbf{P} & \textbf{R} & \textbf{F1} \\ \hline
RoBERTa & 0,827 & 0,8264 & 0,8260\\
BERTweet & 0,8382 & 0,8376 & 0,8373 \\
FlairEmbeddings & 0.7066 & 0.7064 & 0.7061 \\
\hline
\end{tabular}
\caption{These models were trained only with FALSE and TRUE statements. The scores indicate the \textbf{P}recision, the \textbf{R}ecall and the \textbf{F1}-Score}
\label{table:binclass}
\end{table} % results Table binary

\subsubsection{Multiclass Classification}
We also trained all three models on the four existing classes. Due to the higher classification complexity all validation scores (shown in Table \ref{table:multiclass}) resulted about 10-20\% worse than for the binary classification. It is worth to mention, that RoBERTa performed slightly better than BERTweet, contrary to the less complex classification in Section \ref{sec:binclass}. FlairEmbeddings' scores were affected the most and dropped to ~50\%.
\begin{table}[ht!]
\centering
\begin{tabular}{llll}
\hline \textbf{Model} & \textbf{P} & \textbf{R} & \textbf{F1} \\ \hline
RoBERTa & 0,7186  &  0,7141  &  0,7154\\
BERTweet & 0,7149  &  0,7140  &  0,7119 \\
FlairEmbeddings & 0,5298 &  0,5055  &  0,5026 \\
\hline
\end{tabular}
\caption{These models were trained only with all existing classes. The scores indicate the \textbf{P}recision, the \textbf{R}ecall and the \textbf{F1}-Score}
\label{table:multiclass}
\end{table} % results table multi

\subsection{Adversarial Attacks}
The next step is to apply adverserial attacks on the dataset using TextAttack. Approximately 40 FALSE statements were sampled from the dataset and attacked using the TextAttack recipes. Due to the poor multiclass classification performance we applied all attacks on the binary-trained models. The results can be found in Table \ref{bert-attacks-table} for BERTweet, in Table \ref{roberta-attacks-table} for RoBERTa and in Table \ref{flair-attacks-table} for Flair. The tables contain the percentage of sentences predicted as Successful, Failed or Skipped. Table \ref{reduction-feng-table} shows the percentages obtained for InputReductionFeng for all three models and the mean value of score improvement over all sentences for each model. To see the distribution of word level and character level attacks, Table \ref{mean-success-score} was generated. It contains the mean value of the Success percentage that can be found in tables \ref{bert-attacks-table}, \ref{roberta-attacks-table} and \ref{flair-attacks-table}. \\

\begin{table}[ht!]
\centering
\begin{tabular}{llll}
\hline \textbf{Recipe}  &\textbf{Success} &  \textbf{Fail} & \textbf{Skipped} \\ 
&(\%)&(\%)&(\%)\\ 
\hline
IGAWang  & 90 & 0 & 10\\
TextFoolerJin2019  & 85 & 0 & 15\\
DeepWordBug  & 77.5 & 7.5 & 15\\
PWWSRen2019  & 77.5 & 7.5 & 15 \\
PSOZang  & 67.5 & 17.5 & 15\\
BAEGarg2019  & 67.5 & 30 & 2.5  \\
TextBugger  & 57 & 28 & 15 \\
Pruthi  & 28 & 57 & 15 \\
CheckList2020  & 20 & 75 & 5 \\
\hline
\end{tabular}
\caption{\label{bert-attacks-table}Attack results for BERTweet. }
\end{table}

 % results table bertweet
\begin{table}[ht!]
\centering
\begin{tabular}{llll}
\hline \textbf{Recipe}  &\textbf{Success} &  \textbf{Fail} & \textbf{Skipped}  \\ 
&(\%)&(\%)&(\%)\\ 
\hline
TextFoolerJin2019  & 92.5 & 5 & 2.5 \\
DeepWordBug  & 87.5 & 10 & 2.5\\
PWWSRen2019  & 77.5 & 20 & 2.5 \\
TextBugger  & 75 & 22.5 & 2.5 \\
IGAWang  & 70 & 10 & 20 \\
PSOZang  & 70 & 20 & 10 \\
BAEGarg2019  & 60 & 37.5 & 2.5 \\
Pruthi  & 37.5 & 60& 2.5 \\
CheckList2020  & 7.5 & 90 & 2.5 \\
\hline
\end{tabular}
\caption{\label{roberta-attacks-table}Attack results for Roberta. }
\end{table}

 % results table RoBERTA

BERTweet and RoBERTa both use document embeddings. They show similar results. For BERTweet the word level attack IGAWang had the highest success rate with 90\%. CheckList2020 was the least successful with 20\%. For RoBERTa the word level attack TextFoolerJin2019 was the most successful with 92.5\% and Checklist2020 the least successful with 7.5\%. With the exception of IGAWang the recipes show the same order in success rates. The character level attack DeepWordBug seems to be very successful for these two model, ranking place 2 for RoBERTa and 3 for BERTweet. This is surprising since in the overall ranking in Table \ref{mean-success-score} it only achieved place 7. This might be an indication that document embeddings are more vulnerable to character level attacks than word embeddings.

\begin{table}[ht!]
\centering
\begin{tabular}{llll}
\hline \textbf{Recipe}& \textbf{Success} & \textbf{Fail} & \textbf{Skipped} \\ 
&(\%)&(\%)&(\%)\\ 
\hline
TextFoolerJin2019   & 66 & 6 & 28\\
BAEGarg2019  & 60 & 18 & 22 \\
IGAWang & 50 & 20 & 30\\
PWWSRen2019  & 50 & 22 & 28\\
PSOZang  & 46 & 26 & 28\\
TextBugger & 37 & 26& 37\\
DeepWordBug & 24 & 48  & 28\\
Pruthi  & 18 & 54  & 28\\
CheckList2020 & 2 & 76 & 22 \\
\hline
\end{tabular}
\caption{Attack results for Flair Embeddings.}
\label{flair-attacks-table}
\end{table}

% results table flair

In comparison to BERT-based models Flair classified a lot more inputs wrongly, which is depicted in the higher percentage of skipped statements in Table \ref{flair-attacks-table}. Nevertheless, it proved to be a lot less vulnerable towards adverserial attacks considering that the best performing recipe TextFoolerJin2019 only reached a success rate of 66\% (vs 90\% and 92.5\%). Similar to the previous models, CheckList2020 performed worse, but this time with a success rate of only 2\%. Both pure character level attack recipes failed most of their attacks with 48\% (DeepWordBug) and 54\% (Pruthi) fail rate. Overall, it appears that the contextualization used by Flair makes the model a lot robuster towards the used word and character level based TextAttack recipes.

% results Table InputReductionFeng
%\textit{InputReductionFeng} attacks have different result outputs and were added to a separate table (please refer to section 4.5): 
\begin{table}[ht!]
\centering
\begin{tabular}{llll}
\hline \textbf{Model} & \textbf{Maximized} & \textbf{Skipped} &\textbf{Score}\\ 
&(\%)&(\%)&\textuparrow $\phi$ \\
\hline
BERTweet & 83 & 17&+0.65 \\
Roberta & 81 & 19&+0.63\\
Flair & 75 & 25& +0.64 \\
\hline
\end{tabular}
\caption{\label{reduction-feng-table}Attack results for InputReductionFeng. The result outputs of this attack are described in Section \ref{sec:textattack}.}
\end{table}
BERTweet seems to be the most vulnerable to InputReductionFeng attacks but the difference in score increasing is not very high (+-0.01). Overall it seems that these attacks show similar results over all models.

% results Table success rate over all models
\begin{table}[ht!]
\centering
\begin{tabular}{llll}
\hline \textbf{Recipe}  &\textbf{Success} & \textbf{Level} \\ 
&(\%)&\\ 
\hline
TextFoolerJin2019  & 81.17 & word\\
IGAWang  & 70 & word\\
PWWSRen2019  & 77.5  & word \\
PSOZang  & 68.33  & word\\
BAEGarg2019  & 67.5 & word \& character \\
DeepWordBug  & 63  & character\\
TextBugger  & 62.5 &  word \& character\\
Pruthi  & 27.83 & character\\
CheckList2020  & 9.83  & word \\
\hline
\end{tabular}
\caption{\label{mean-success-score}Mean Success Rates (all models).}
\end{table}

Table 8 shows that the 4 top ranking attacks are word level attacks. It seems that the models are more vulnerable to word level attacks than to character level or mixed attacks (character \& word). \\

Finally the Success Rate, for every model was calculated using formula \ref{eq:1}: 

\begin{equation}
\label{eq:1}
S_r = { \sum_{i=1}^{a} { s_{i} \over  s_{i} + f_{i}} \over a}
\end{equation}
with $S_r$ being the success rate $s$ being the number of successful attacks,
$f$ being the number of failed attacks and $a$ being number of attacks recipes.
\begin{table}[ht!]
\centering
\begin{tabular}{ll}
\hline \textbf{Model}  &\textbf{$S_r$} \\ 
&(\%)\\ 
\hline
BERTweet  & 72,45 \\
Roberta  & 68,22 \\
Flair  & 54,77 \\
\hline
\textbf{Total Average} & \textbf{65,15}\% \\
\end{tabular}
\caption{\label{sucess-rates}Success Rates. }
\end{table}

The skipped statements were not taken into the calculation, as they are dependent on the model training and not on TextAttack. The skipped values show which statements were not predicted correctly by the model in the first place. Thus, they are depended on the model and reflect models accuracy. For the success rates we wanted to see TextAttack's efficiency on statements that the model would correctly classify as fake news.  Our results for the Success Rates underline the results seen in tables \ref{bert-attacks-table}, \ref{roberta-attacks-table} and \ref{flair-attacks-table}, showing that Flair is less vulnerable to the attacks with S$_r$=54,77\% than the other two models with S$_r$= $\sim$70\%. As an conclusion this gives us an Total Success Rate of 65,15\% for adversarial attacks on fake news detection using TextAttack.

% CONCLUSION
\section{Conclusion}
\label{sec:conclusion}
This paper aimed to answers the question: how vulnerable are automatic fake news detection to adversarial attacks? We tested this by checking if automated augmentation of fake news sentences (FALSE statements) will lead to TRUE classifications. This would allow them to bypass the fake news detection mechanisms. Our results show that using the python library TextAttack allows automated changing of classification for 65,15\% of the sentences. Flair, the only model using word-level embedding (contextual string embeddings), seems less vulnerable to attacks with a Success Rate of 54,77\%. The other two models using document embeddings show 72,45\% (BERTweet) and 68,22\% (Roberta). Furthermore, word-level swaps seem to be slightly more successful with an average of 76,87\% compared to character level or mixed swaps with average 55,21\%. Consequently, the models are more vulnerable to attacks using semantically correct sentences with changed meaning than to attacks using typos. Overall it seems that it is possible to bypass the classifier with these attacks. However, our results do not consider that a human will be able to see obvious spelling mistakes in sentences. Furthermore, a human will have a higher accuracy of recognizing unfitting words in sentences. Looking at the augmented sentences we think that many of them will be recognized as FALSE (\textit{TrOmp} \textit{hsut} down American  \textit{airportA} on 4  \textit{Jul} \textit{201B} or Hollywood Action Star \textit{Christelle} Chan Dead).
\\
As a conclusion of these results we think that using the policy initiatives blocking and deprioritization, should be avoided if possible, as these methods don't completely remove the fake news statements from the users feeds. This makes it easier to target these statements with automated attacks as shown in our research. A scenario would be to attack these sentences until they are unblocked or re-prioritized in a users feed. This would lead to dangerous spreading of Fake News in social networks. %damdamdamdam

\section*{Acknowledgments}
We would like to thank Dr. Vinicius Woloszyn for his support and supervision of the project.

\bibliography{anthology,eacl2021,bibliography}

\begin{thebibliography}{31}
\expandafter\ifx\csname natexlab\endcsname\relax\def\natexlab#1{#1}\fi

\bibitem[{Akbik et~al.(2018)Akbik, Blythe, and Vollgraf}]{flairembeddings}
Alan Akbik, Duncan Blythe, and Roland Vollgraf. 2018.
\newblock Contextual string embeddings for sequence labeling.
\newblock In \emph{{COLING} 2018, 27th International Conference on
  Computational Linguistics}, pages 1638--1649.

\bibitem[{Chakraborty et~al.(2016)Chakraborty, Paranjape, Kakarla, and
  Ganguly}]{chakraborty2016stop}
Abhijnan Chakraborty, Bhargavi Paranjape, Sourya Kakarla, and Niloy Ganguly.
  2016.
\newblock \href {http://arxiv.org/abs/1610.09786} {Stop clickbait: Detecting
  and preventing clickbaits in online news media}.

\bibitem[{Comission(2018)}]{expert_group}
European Comission. 2018.
\newblock \emph{Final report of the High Level Expert Group on Fake News and
  Online Disinformation}.
\newblock Publications Office of the European Union.

\bibitem[{Devlin et~al.(2019)Devlin, Chang, Lee, and Toutanova}]{bert}
Jacob Devlin, Ming-Wei Chang, Kenton Lee, and Kristina Toutanova. 2019.
\newblock \href {http://arxiv.org/abs/1810.04805} {Bert: Pre-training of deep
  bidirectional transformers for language understanding}.

\bibitem[{Feng et~al.(2018)Feng, Wallace, au2, Iyyer, Rodriguez, and
  Boyd-Graber}]{feng2018pathologies}
Shi Feng, Eric Wallace, Alvin Grissom~II au2, Mohit Iyyer, Pedro Rodriguez, and
  Jordan Boyd-Graber. 2018.
\newblock \href {http://arxiv.org/abs/1804.07781} {Pathologies of neural models
  make interpretations difficult}.

\bibitem[{Gao et~al.(2018)Gao, Lanchantin, Soffa, and Qi}]{gao2018blackbox}
Ji~Gao, Jack Lanchantin, Mary~Lou Soffa, and Yanjun Qi. 2018.
\newblock \href {http://arxiv.org/abs/1801.04354} {Black-box generation of
  adversarial text sequences to evade deep learning classifiers}.

\bibitem[{Garg and Ramakrishnan(2020)}]{garg2020bae}
Siddhant Garg and Goutham Ramakrishnan. 2020.
\newblock \href {http://arxiv.org/abs/2004.01970} {Bae: Bert-based adversarial
  examples for text classification}.

\bibitem[{Ghanem et~al.(2021)Ghanem, Ponzetto, Rosso, and Rangel}]{q1_fakeflow}
Bilal Ghanem, Simone~Paolo Ponzetto, Paolo Rosso, and Francisco Rangel. 2021.
\newblock Fakeflow: Fake news detection by modeling the flow of affective
  information.
\newblock \emph{arXiv preprint arXiv:2101.09810}.

\bibitem[{Gokaslan and Cohen(2019)}]{openwebtext}
Aaron Gokaslan and Vanya Cohen. 2019.
\newblock Openwebtext corpus.
\newblock \url{http://Skylion007.github.io/OpenWebTextCorpus}.

\bibitem[{Jin et~al.(2020)Jin, Jin, Zhou, and Szolovits}]{jin2020bert}
Di~Jin, Zhijing Jin, Joey~Tianyi Zhou, and Peter Szolovits. 2020.
\newblock \href {http://arxiv.org/abs/1907.11932} {Is bert really robust? a
  strong baseline for natural language attack on text classification and
  entailment}.

\bibitem[{Li et~al.(2019)Li, Ji, Du, Li, and Wang}]{Li_2019}
Jinfeng Li, Shouling Ji, Tianyu Du, Bo~Li, and Ting Wang. 2019.
\newblock \href {https://doi.org/10.14722/ndss.2019.23138} {Textbugger:
  Generating adversarial text against real-world applications}.
\newblock \emph{Proceedings 2019 Network and Distributed System Security
  Symposium}.

\bibitem[{Liu et~al.(2019)Liu, Ott, Goyal, Du, Joshi, Chen, Levy, Lewis,
  Zettlemoyer, and Stoyanov}]{roberta}
Yinhan Liu, Myle Ott, Naman Goyal, Jingfei Du, Mandar Joshi, Danqi Chen, Omer
  Levy, Mike Lewis, Luke Zettlemoyer, and Veselin Stoyanov. 2019.
\newblock \href {http://arxiv.org/abs/1907.11692} {Roberta: A robustly
  optimized bert pretraining approach}.

\bibitem[{Marsden and Meyer(2019)}]{euregulation}
Chris Marsden and Trisha Meyer. 2019.
\newblock \emph{Regulating disinformation with artificial intelligence: effects
  of disinformation initiatives on freedom of expression and media pluralism}.
\newblock European Parliament.

\bibitem[{Morris et~al.(2020)Morris, Lifland, Yoo, Grigsby, Jin, and
  Qi}]{textattack}
John Morris, Eli Lifland, Jin~Yong Yoo, Jake Grigsby, Di~Jin, and Yanjun Qi.
  2020.
\newblock Textattack: A framework for adversarial attacks, data augmentation,
  and adversarial training in nlp.
\newblock In \emph{Proceedings of the 2020 Conference on Empirical Methods in
  Natural Language Processing: System Demonstrations}, pages 119--126.

\bibitem[{Nagel(2016)}]{ccnews}
Sebastian Nagel. 2016.
\newblock Cc-news.
\newblock \url{https://commoncrawl.org/2016/10/news-dataset-available/}.

\bibitem[{Nguyen et~al.(2020)Nguyen, Vu, and Nguyen}]{bertweet}
Dat~Quoc Nguyen, Thanh Vu, and Anh~Tuan Nguyen. 2020.
\newblock {BERTweet: A pre-trained language model for English Tweets}.
\newblock In \emph{Proceedings of the 2020 Conference on Empirical Methods in
  Natural Language Processing: System Demonstrations}.

\bibitem[{Pruthi et~al.(2019)Pruthi, Dhingra, and Lipton}]{pruthi2019combating}
Danish Pruthi, Bhuwan Dhingra, and Zachary~C. Lipton. 2019.
\newblock \href {http://arxiv.org/abs/1905.11268} {Combating adversarial
  misspellings with robust word recognition}.

\bibitem[{Ren et~al.(2019)Ren, Deng, He, and Che}]{ren-etal-2019-generating}
Shuhuai Ren, Yihe Deng, Kun He, and Wanxiang Che. 2019.
\newblock \href {https://doi.org/10.18653/v1/P19-1103} {Generating natural
  language adversarial examples through probability weighted word saliency}.
\newblock In \emph{Proceedings of the 57th Annual Meeting of the Association
  for Computational Linguistics}, pages 1085--1097, Florence, Italy.
  Association for Computational Linguistics.

\bibitem[{Ribeiro et~al.(2020)Ribeiro, Wu, Guestrin, and
  Singh}]{ribeiro2020accuracy}
Marco~Tulio Ribeiro, Tongshuang Wu, Carlos Guestrin, and Sameer Singh. 2020.
\newblock \href {http://arxiv.org/abs/2005.04118} {Beyond accuracy: Behavioral
  testing of nlp models with checklist}.

\bibitem[{Schneider(2021)}]{githubcode}
Filla Schneider, Koenders. 2021.
\newblock Code.
\newblock
  \url{https://github.com/nicolaischneider/FakeNewsDetectionVulnerability}.

\bibitem[{Shu et~al.(2019)Shu, Cui, Wang, Lee, and Liu}]{q3_defend}
Kai Shu, Limeng Cui, Suhang Wang, Dongwon Lee, and Huan Liu. 2019.
\newblock defend: Explainable fake news detection.
\newblock In \emph{Proceedings of the 25th ACM SIGKDD international conference
  on knowledge discovery \& data mining}, pages 395--405.

\bibitem[{Tchechmedjiev et~al.(2019)Tchechmedjiev, Fafalios, Boland, Gasquet,
  Zloch, Zapilko, Dietze, and Todorov}]{tchechmedjiev2019claimskg}
Andon Tchechmedjiev, Pavlos Fafalios, Katarina Boland, Malo Gasquet,
  Matth{\"a}us Zloch, Benjamin Zapilko, Stefan Dietze, and Konstantin Todorov.
  2019.
\newblock Claimskg: a knowledge graph of fact-checked claims.
\newblock In \emph{International Semantic Web Conference}, pages 309--324.
  Springer.

\bibitem[{Trinh and Le(2019)}]{storiesdataset}
Trieu~H. Trinh and Quoc~V. Le. 2019.
\newblock \href {http://arxiv.org/abs/1806.02847} {A simple method for
  commonsense reasoning}.

\bibitem[{Wang et~al.(2021)Wang, Jin, Yang, and He}]{wang2021natural}
Xiaosen Wang, Hao Jin, Yichen Yang, and Kun He. 2021.
\newblock \href {http://arxiv.org/abs/1909.06723} {Natural language adversarial
  defense through synonym encoding}.

\bibitem[{Woloszyn et~al.(2021)Woloszyn, Cortes, Amantea, Schmitt, Barone, and
  M{\"o}ller}]{q2_claimreview}
Vinicius Woloszyn, Eduardo~G Cortes, Rafael Amantea, Vera Schmitt, Dante~AC
  Barone, and Sebastian M{\"o}ller. 2021.
\newblock Towards a novel benchmark for automatic generation of claimreview
  markup.
\newblock In \emph{13th ACM Web Science Conference 2021}, pages 29--35.

\bibitem[{Woloszyn et~al.(2020)Woloszyn, Schaeffer, Boniatti, Cortes, Mohtaj,
  and M{\"o}ller}]{woloszyn2020untrue}
Vinicius Woloszyn, Felipe Schaeffer, Beliza Boniatti, Eduardo Cortes, Salar
  Mohtaj, and Sebastian M{\"o}ller. 2020.
\newblock Untrue.news: A new search engine for fake stories.
\newblock \emph{arXiv preprint arXiv:2002.06585}.

\bibitem[{Zang et~al.(2020)Zang, Qi, Yang, Liu, Zhang, Liu, and
  Sun}]{Zang_2020}
Yuan Zang, Fanchao Qi, Chenghao Yang, Zhiyuan Liu, Meng Zhang, Qun Liu, and
  Maosong Sun. 2020.
\newblock \href {https://doi.org/10.18653/v1/2020.acl-main.540} {Word-level
  textual adversarial attacking as combinatorial optimization}.
\newblock \emph{Proceedings of the 58th Annual Meeting of the Association for
  Computational Linguistics}.

\bibitem[{Zhixuan~Zhou and Hsu(2021)}]{via_NLP}
Meghana Moorthy~Bhat Zhixuan~Zhou, Huankang~Guan and Justin Hsu. 2021.
\newblock Fake news detection via nlp is vulnerable to adversarial attacks.
\newblock \emph{arXiv preprint arXiv:2101.09810}.

\bibitem[{Zhou and Zafarani(2019)}]{surveyzhouzafarani}
Xinyi Zhou and Reza Zafarani. 2019.
\newblock \emph{Fake news: A survey of research, detection methods, and
  opportunities}.
\newblock arXiv preprint arXiv:1901.09657.

\bibitem[{Zhou et~al.(2019)Zhou, Guan, Bhat, and Hsu}]{zhou2019fake}
Zhixuan Zhou, Huankang Guan, Meghana~Moorthy Bhat, and Justin Hsu. 2019.
\newblock Fake news detection via nlp is vulnerable to adversarial attacks.
\newblock \emph{arXiv preprint arXiv:1901.09657}.

\bibitem[{Zhu et~al.(2015)Zhu, Kiros, Zemel, Salakhutdinov, Urtasun, Torralba,
  and Fidler}]{bookcorpus}
Yukun Zhu, Ryan Kiros, Richard Zemel, Ruslan Salakhutdinov, Raquel Urtasun,
  Antonio Torralba, and Sanja Fidler. 2015.
\newblock \href {http://arxiv.org/abs/1506.06724} {Aligning books and movies:
  Towards story-like visual explanations by watching movies and reading books}.

\end{thebibliography}
\bibliographystyle{acl_natbib}

\end{document}